\documentclass[10pt,twocolumn,letterpaper]{article}

\usepackage{avss}
\usepackage{times}
\usepackage{setspace}
\usepackage{epsfig}
\usepackage{graphicx}
\usepackage{multirow}
\usepackage{amsmath}
\usepackage{amssymb}
\usepackage{array}
\usepackage{balance}

\def\Hline{\noalign{\hrule height 4\arrayrulewidth}}
\newcolumntype{V}{>{$\vcenter\bgroup\hbox\bgroup}c<{\egroup\egroup$}}

\avssfinalcopy

\def\avssPaperID{1} % *** Enter the AVSS Paper ID here

\graphicspath{{./imgs/}}

\ifavssfinal\fi
\begin{document}

%%%%%%%%% TITLE
\title{Automatic Vehicle Tracking and Recognition from Aerial Image Sequences}

\author{Ognjen Arandjelovi\'c\\
School of Computer Science\\University of St Andrews\\St Andrews\\Fife KY16 9SX\\Scotland\\\small E-mail: \texttt{ognjen.arandjelovic@gmail.com}}

\maketitle

\begin{abstract}
    This paper addresses the problem of automated vehicle tracking and recognition from aerial image sequences. Motivated by its successes in the existing literature
    we focus on the use of linear appearance subspaces to describe multi-view object appearance and highlight the challenges involved in their application as a part of a practical system. A working solution which includes steps for data extraction and normalization is described. In experiments on real-world data the proposed methodology achieved promising results with a high correct recognition rate and few, meaningful errors (type II errors whereby \emph{genuinely similar} targets are sometimes being confused with one another). Directions for future research and possible improvements of the proposed method are discussed.
\end{abstract}

%%%%%%%%% BODY TEXT
\section{Introduction}
In this paper we address the problem of automatic recognition of ground objects (mainly vehicles) from image sequences acquired from unmanned aircraft. This is a very challenging recognition scenario in which viewpoint and illumination can vary greatly, occlusions and background clutter are common, and data is of low quality. Generally speaking, target recognition systems comprise two distinct tasks, that of (i) representing object appearance and of subsequent (ii) representation matching. Both of these tasks are pervasive across different object recognition and matching problems, and have attracted a significant amount of research attention in the computer vision community~\cite{AggaRoth2002,AhmaKitt2002,BelhKrie1998}. This interest has particularly intensified in recent years, after significant advances towards practically viable systems have been made~\cite{Aran2012f,AranZiss2011,BelhKrie1998,SiviRussEfro+2005}.

The most prominent group approaches are local descriptor-based. Methods of this group employ descriptors in a sparse fashion by focusing on a set of automatically localized interest points~\cite{BregXianGong2012,HeikPietSchm2009}. When the number of detected interest points is large this approach can achieve impressive robustness to partial occlusion and pose~\cite{Lowe2004}. However, a serious limitation of this approach is that it cannot deal well with untextured objects (sometimes referred to as smooth objects) which by their very nature do not exhibit appearance which results in a larger number of consistently well-localized interest points. This limitation has recently attracted increased research attention; shape-based approaches using boundary appearance features~\cite{AranZiss2011} or pure shape~\cite{Aran2012i,Aran2012f} have demonstrated promising results on databases of objects which have distinct shapes. Other approaches include part-based methods, suitable to the recognition of articulated objects with distinctly recognizable parts (such as a face which can be seen as comprising two `eye parts', a `mouth part' etc. which vary in mutual configuration depending on the person's pose and facial expression)~\cite{FelzHutt2005}. A summary of different representation and matching techniques dominant in the existing literature is shown in Tables~\ref{t:representation} and~\ref{t:matching}.

There are two key conceptual contributions we make to the current state-of-the-art. Firstly most existing methods address the problem at hand using individual images. In contrast in this paper the focus is on recognition from image sequences. In other words a sequence of frames/images of an unknown, query object is matched against a database of sequences of known, database targets. This problem setting is of an increasing significance considering the ease with which in our application image sequences can be acquired and stored. Secondly we show how a robust and pose-invariant system can be built by enriching the set of directly acquired exemplars with synthetically generated data, and how the resulting sets can be appropriately described and matched. We first summarize the key ideas and then explain each element of our method in detail in the next section.

\paragraph{Approach summary}
Considering the small scale of the target, model or part-based approaches are usually unsuitable for the problem in question. Notwithstanding its limitations, in this paper we focus on the use of a multi-view model in the form of a set of raw holistic appearance patches. This model motivates increasingly popular manifold-based approaches for matching, which exploit the structure of object's appearance change as viewing parameters are varied. Following its success in related recognition problems and the potential for real-time performance in an incremental learning framework due to its computational efficiency and low storage requirements, here we specifically examine the use of canonical correlation analysis~\cite{AranCipo2013}.

\begin{table}[t]
  \LARGE
  \centering
  \caption{Classification of the most influential object recognition representations.}
  \vspace{3pt}
  \begin{tabular}{l|c}
    \Hline

    \begin{tabular}{l}
      \rotatebox{90}
      {\small \bf Object representations}\\
    \end{tabular}&

    \begin{tabular}{l|l}
      \multirow{2}{*}{\small Global representations} &
         \small Appearance prototypes~\cite{LeCuHuanBott2004} \\
       & \small Model-based~\cite{MianBennOwen2006} \\
      \hline
      \multirow{3}{*}{\small Local representations} &
        \small Pictorial structure~\cite{FelzHutt2005} \\
      & \small Dense local features~\cite{HeikPietSchm2009} \\
      & \small Sparse local features~\cite{Lowe2004} \\
    \end{tabular}\\
    \Hline
  \end{tabular}
  \label{t:representation}
\end{table}

\begin{table}[!t]
  \LARGE
  \centering
  \caption{Classification of the most influential representation matching approaches.}
  \vspace{3pt}
  \begin{tabular}{l|c}
    \Hline

    \begin{tabular}{l}
      \rotatebox{90}
      {\small \bf Model matching}\\
    \end{tabular}&

    \begin{tabular}{l|l}
      \multirow{2}{*}{\small Single-view} &
         \small Euclidean distance~\cite{LeCuHuanBott2004} \\
       & \small Cosine distance~\cite{ChenErWu2006} \\
      \hline
      \multirow{3}{*}{\small Multi-view} &
        \small Single-view aggregation-based~\cite{FerrTuytVanG2004a} \\
      & \small Probability density-based~\cite{ShakFishDarr2002} \\
      & \small Embedded manifold-based~\cite{ShasLeviAvid2002} \\
    \end{tabular}\\
    \Hline
  \end{tabular}
  \label{t:matching}
\end{table}

\section{Method details}
In this section we explain each of the constituent elements of the proposed algorithm. We start by summarizing the motivation and theory underlying the approach we adopt for matching sets of exemplars. Then we explain how exemplar data is extracted from raw imagery: Section~\ref{ss:tracking} describes how a detected target is tracked through a sequence of images while possibly undergoing pose change, while Section~\ref{s:synth} explains how viewpoint invariance is achieved by enriching the explicitly extracted set of exemplars with additional, synthetically generated images.

\subsection{Overview of the baseline approach}\label{s:cca} For classification, canonical correlation analysis \cite{Hote1936} is usually employed by computing canonical correlations (i.e.\ the cosines of principal angles) between linear manifolds \cite{FukuYama2003,NishYama2006,WolfShas2003}. Canonical correlations, $0 \leq \theta_1 \leq \ldots \leq \theta_d \leq (\pi/2)$ between any two $d$-dimensional linear manifolds or hyperplanes $\mathcal{L}_1$ and $\mathcal{L}_2$, are uniquely defined as~\cite{BachJord2002}:
\begin{equation}\label{Eqn: Principal Angles Definition 1}
    \cos \theta_i=\max_{\mathbf{u}_i\in \mathcal{L}_1}\max_{\mathbf{v}_i\in \mathcal{L}_2}
                \mathbf{u}_i^T \mathbf{v}_i
\end{equation}
\noindent subject to $ \mathbf{u}_i^T \mathbf{u}_i = \mathbf{v}_i^T
\mathbf{v}_i = 1,~
   \mathbf{u}_i^T \mathbf{u}_j = \mathbf{v}_i^T \mathbf{v}_j = 0,~
   i \neq j $.
The solution can be obtained by applying the Singular Value Decomposition (SVD)~\cite{BjorGolu1973}, whose complexity is $O(d^3)$ where $d$ is the dimensionality of the manifolds ($d$ is typically small). Each image set is represented by a linear manifold and the angles between two low-dimensional manifolds are exploited as a similarity measure between two image sets. The canonical vectors in each pair are visually similar despite the large changes of object pose. The first pair of canonical vectors corresponds to the most correlated vectors, each of which is spanned by any linear combination of the respective image set. The next pairs of canonical vectors represent the directions of the next most similar data variations of the two sets in other dimensions. CCA effectively finds common modes (e.g.\ target object pose) of two image sets.

\subsection{Data extraction}\label{s:extraction} 
\begin{figure}[!th]
  \centering
  \includegraphics[width=0.48\textwidth]{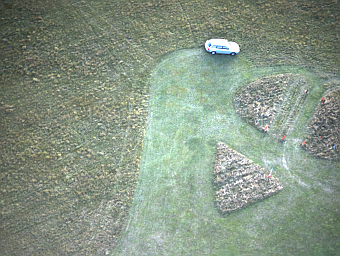}
  \vspace{0pt}
  \caption{Raw input frame showing a poorly localized target by the aircraft's on-board system.}
          \label{f:loc1}
\end{figure}

In designing a practical object recognition system it is of crucial importance to appreciate that the actual recognition algorithm operates in the context of the preceding data acquisition and extraction stages. Specifically, in the task of target recognition from aerial images acquired using unmanned aircraft, the target needs to be located and tracked. On the lowest level, this is done by the on-board control system which employs the GPS coordinates of the target and the plane, and a gyroscope. Difficulties are already introduced here -- the target moves widely across the image, as shown in Figure~\ref{f:loc1}, often even entirely disappearing from the view. Given that vehicles appear in aerial images mostly as smooth object, resulting in few stably detectable keypoints, we found local-feature-based tracking~\cite{MartAran2010} unreliable. Instead we propose an alternative solution which comprises two steps. Firstly, the problem of initializing tracking by detection is solved by registering consecutive image frames globally. This is readily achieved because video sequences of interest in this paper are mostly comprised of static backgrounds. Hence in the context of registration of frames the target can be considered an outlier (in that it moves from a frame to frame). After consecutive frames are registered, the target itself is easily detected by performing simple background subtraction. To account for noise, morphological image processing is performed to remove spurious regions of frame-to-frame difference, with the correct region reliably being detected as the one with the largest contiguous area over which the aforementioned difference exceeds a threshold.

\subsubsection{Target tracking}\label{ss:tracking} Following the localization of the target, we track it until it is no longer entirely visible. For this we employ the well known Lucas-Kanade tracker~\cite{Aran2011a,BakeMatt2004}, with 6 affine degrees of freedom. In this algorithm, at each image-to-image transition in a sequence, the generalized position vector is iteratively updated by minimization of the error (i.e.\ appearance difference in Euclidean distance sense) between the region of an image at that position and the warped template of the target from the preceding image in the sequence. In more detail, given the appearance of the target $I_i(x,y)$ in the $i$-th image in a sequence, the corresponding region of interest in the subsequent frame $I_{i+1}$ is found by minimizing:
\begin{align}
  \sum_{(x,y)\in\mathcal{R}} \left[ I_{i+1}(\mathcal{W}(x,y;\mathbf{p}))-I_i(x,y) \right],
\end{align}
where $\mathcal{R}$ is the quadrilateral region of interest specifying the target in $I_i$, $\mathcal{W}$ the warping function (an affine warp in our case), and $\mathbf{p}$ the warp parameters. The optimal $\mathbf{p}$ is found through iterative descent by linearizing $I_{i+1}(\mathcal{W}(x,y;\mathbf{p}))$ using Taylor expansion, giving:
\begin{align}
  \sum_{(x,y)\in\mathcal{R}} \left[ I_{i+1}(\mathcal{W}(x,y;\mathbf{p}_j))+\nabla I \frac{\partial\mathcal{W}}{\partial p}\Delta \mathbf{p}-I_i(x,y) \right],
\end{align}
where $\mathbf{p}_0$ are the initial warp parameters, and $\mathbf{p}_1 \ldots \mathbf{p}_j$ the iterative refinement sequence. We found this approach to work well on our data set and expect a similar level of performance on imagery acquired under similar conditions (elevation and angle to the target).

\subsubsection{Appearance set generation}\label{s:synth} Ideally, the target recognition system receives views of the target across 360$^{\circ}$ range, obtained by a circular reconnaissance manoeuver over the target. However, as a consequence of the difficulties involved in the camera's control system locking onto the target (see
Section~\ref{s:extraction}), the range of views obtained from each flyover is incomplete. Furthermore, in the experiments reported in this paper, it was decided to use each tracking burst (from the initial detection of the target until the target is lost as described in Section~\ref{ss:tracking}) as a single training set, thus
restricting the range of views available for training. The reason for this lies in the problem of concatenating aforementioned tracking bursts into one data set without introducing training data artefacts. These include, for example, repeated views of the target or differing aeroplane viewpoint angle due to multiple flyovers.

The described fragmentation of input video sequences is a serious problem as the baseline method of interest offers no invariance in this regard. As described in Section~\ref{s:cca}, for canonical correlations to extract meaningful similarity between two data sets, a common form of variation must exist. While robust to the presence of dissimilar data, whether due to different viewing conditions or noise, true rotational or view invariance is not inherent in this approach.

\begin{figure}[!t]
  \centering
  \includegraphics[width=0.45\textwidth]{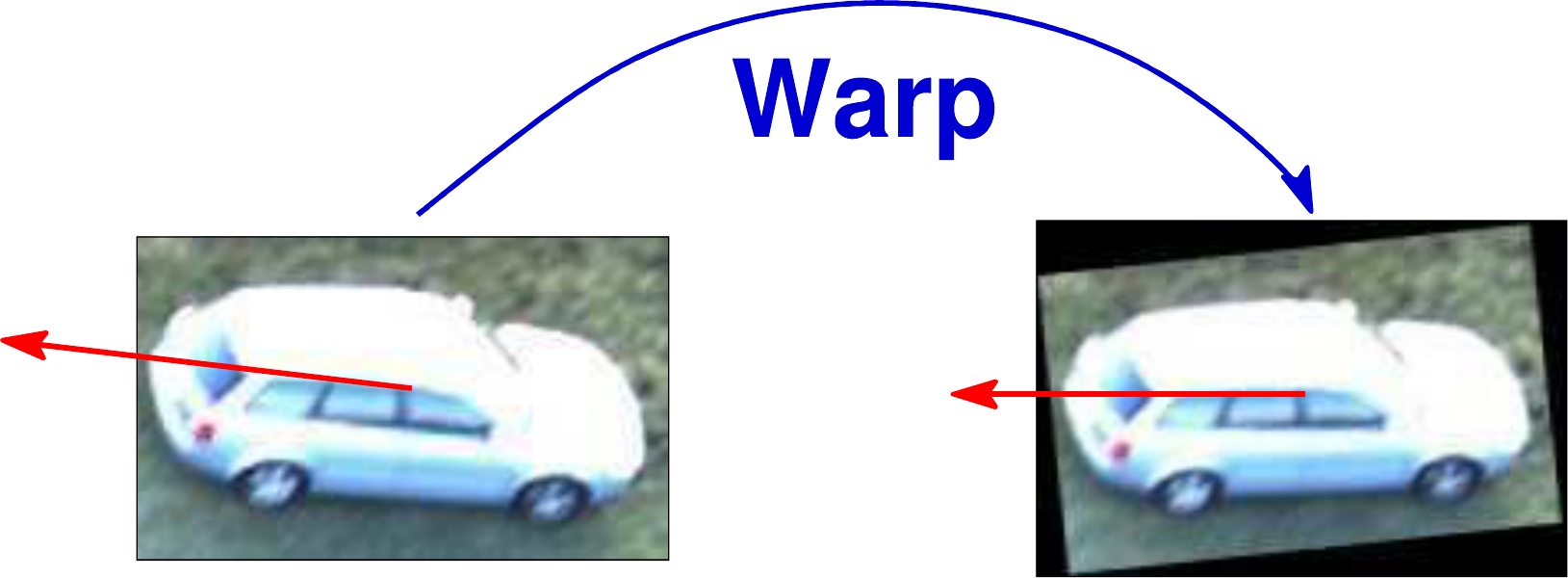}
  \caption{The first step of our data processing and normalization involves ``re-wrapping'' each tracked patch to fit the initial pose.}
          \label{f:warp}
\end{figure}

\begin{figure}[!t]
  \centering
  \begin{tabular}{VV}
    \includegraphics[scale=0.6]{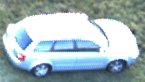}&
    \includegraphics[scale=0.55]{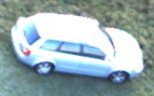}\\
    \footnotesize 0$^{\circ}$ & \footnotesize 10$^{\circ}$ \\
  \end{tabular}
  \begin{tabular}{VVV}
    \includegraphics[scale=0.5]{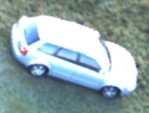}&
    \includegraphics[scale=0.5]{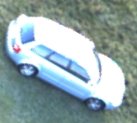}&
    \includegraphics[scale=0.5]{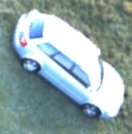}\\
    \footnotesize 20$^{\circ}$ & \footnotesize 30$^{\circ}$ & \footnotesize 40$^{\circ}$\\
  \end{tabular}
  \begin{tabular}{VV}
    \includegraphics[scale=0.5]{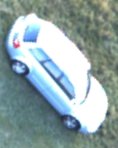}&
    \includegraphics[scale=0.5]{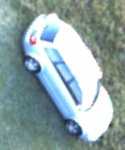}\\
    \footnotesize 50$^{\circ}$ & \footnotesize 60$^{\circ}$ \\
  \end{tabular}
  \caption{Seven synthetically generated views of the target, produced by rotating the
           ``re-wrapped'' patch (see Figure~\ref{f:warp}) across different angles.}
          \label{f:synth}
\end{figure}

Instead of attempting to achieve invariant matching, in this paper we examine an alternative approach whereby invariance is achieved explicitly, by synthetic data augmentation. We summarize the key stages in the proposed method:
\begin{enumerate}
  \item \textbf{Re-warping:} Our tracking approach involves the estimation of pose parameters for the target. To quasi-normalize this view, we ``re-warp'' the target to fit its initial pose (i.e.\ the pose in the first frame as explained in Section~\ref{s:extraction}), as shown in Figure~\ref{f:warp}.
  \item \textbf{Full view generation:} The ``re-warped'' image of the target is now synthetically rotated across 360$^{\circ}$ at 10$^{\circ}$ intervals, as shown in Figure~\ref{f:synth}.
  \item \textbf{Subspace estimation:} Synthetic views from all tracked templates are compiled and used to estimate a linear subspace, which is used as the final representation of target's appearance in the video.
\end{enumerate}

\section{Empirical evaluation}

In the empirical evaluation reported in this paper, we used data acquired during five flights and multiple target flyovers. The details of each session are summarized in Table~\ref{t:flights}.

\begin{table}[!t]
  \centering
  \small
  \renewcommand{\arraystretch}{1.5}
  \caption{A summary of data obtained during the five flights which were used for empirical evaluation.}
  \vspace{3pt}
  \begin{tabular}{llcc}
  \Hline
   \bf Flight & \bf Date & \bf Images & \bf Resolution\\
   \hline
   1 & 28 October 2009  & 28364 & 1360$\times$1024\\
   2 & 06 November 2009 & 34262 & 1360$\times$1024\\
   3 & 10 November 2009 & 17636 & 1360$\times$1024\\
   4 & 12 November 2009 & 14076 & 1360$\times$1024\\
   5 & 17 November 2009 & 21108 & 1360$\times$1024\\
   \Hline
  \end{tabular}
  \label{t:flights}
\end{table}

\begin{figure}[thp]
  \Large
  \centering
  \begin{tabular}{VVV}
    \includegraphics[scale=0.6]{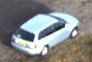}&
    \includegraphics[scale=0.6]{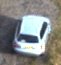}&
    \includegraphics[scale=0.6]{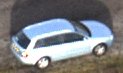}\\
    \multicolumn{3}{c}{\small Car 1}\\[10pt]
    \includegraphics[scale=0.6]{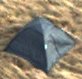}&
    \includegraphics[scale=0.6]{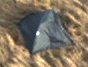}&
    \includegraphics[scale=0.6]{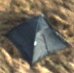}\\
    \multicolumn{3}{c}{\small Tent}\\[10pt]
    \includegraphics[scale=0.48]{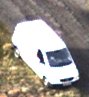}&
    \includegraphics[scale=0.36]{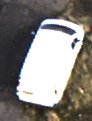}&
    \includegraphics[scale=0.6]{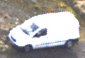}\\
    \multicolumn{3}{c}{\small Van 1}\\[10pt]
    \includegraphics[scale=0.6]{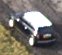}&
    \includegraphics[scale=0.6]{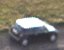}&
    \includegraphics[scale=0.36]{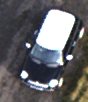}\\
    \multicolumn{3}{c}{\small Car 2}\\[10pt]
    \includegraphics[scale=0.6]{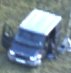}&
    \includegraphics[scale=0.42]{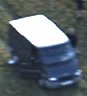}&
    \includegraphics[scale=0.6]{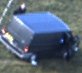}\\
    \multicolumn{3}{c}{\small Van 2}\\[10pt]
    \includegraphics[scale=0.6]{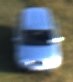}&
    \includegraphics[scale=0.48]{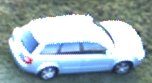}&
    \includegraphics[scale=0.6]{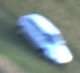}\\
    \multicolumn{3}{c}{\small Car 3}\\[10pt]
  \end{tabular}
  \caption{Six different targets, shown in three different poses each, used to test the proposed algorithm.}
  \label{f:targets}
\end{figure}

To test the proposed recognition system we used six different targets extracted from these five flights. Three views of each and their symbolic names are shown in Figure~\ref{f:targets}. The scale of each target varied throughout input video, depending on its geometry (significantly different for a car and a tent, for example), camera viewpoint, and aeroplane distance (mainly affected by its altitude). The size of the target image patch diagonal lied in the range of 70 to 200 pixels. We normalize tracked target patches by warping them to the uniform scale of $100 \times 100$ pixels.

\subsection{Results} Target recognition was performed by matching a
novel, query data set against training sets of each of the six targets. It was identified as the target that it matched with the highest confidence. The results of rank-1 recognition are summarized using the confusion matrix in Table~\ref{t:cm}. As the confusion matrix illustrates, the proposed method correctly identified the novel target in all but a few cases. An inspection of incorrect target assignments readily shows a clear structure of such errors, targets with genuinely similar appearance being occasionally confused with one another (e.g.\ car~2 and van~2 which are both dark, of similar shape, and only a small difference in size which is not readily apparent from images considering that the camera-target distances are unknown).

\begin{figure}[htb]
  \centering
  \includegraphics[width=0.48\textwidth]{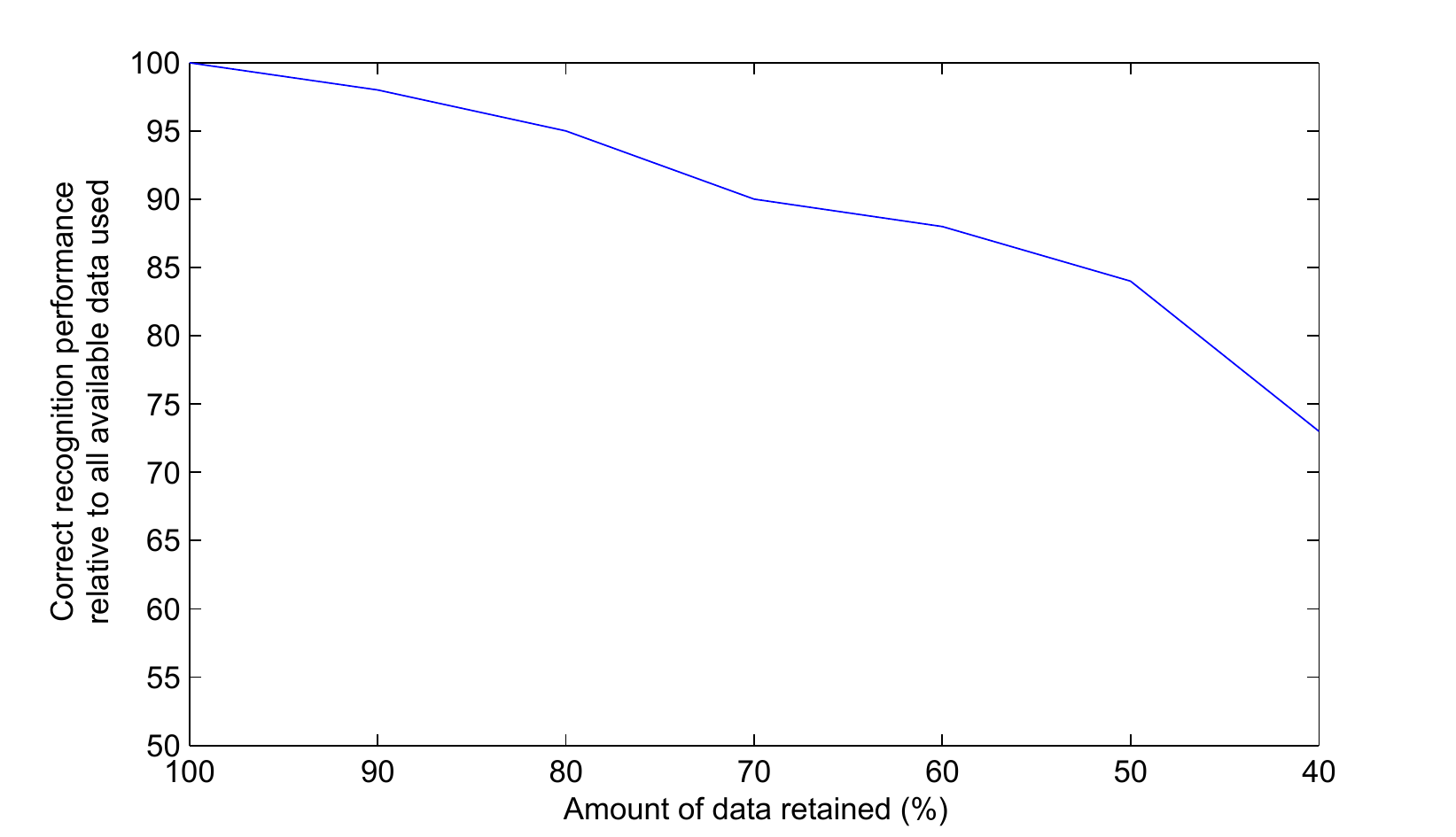}
  \caption{Decay of correct identification rate as the amount of data used for training is reduced. Graceful degradation is demonstrated with a high recognition rate even when 40\% of the data is discarded.}
          \label{f:relative}
\end{figure}

We further tested the sensitivity of our method to the amount of training data. As explained in Section~\ref{s:synth}, in principle even a single image of the target can be used for matching, as a synthetic, compatible 360$^{\circ}$ set is generated from each image. Thus, we explored how the correct recognition rate is affected by gradual removal of an increasing amount of tracked templates. Figure~\ref{f:relative} summarizes the results obtained and shows that the proposed method exhibits slow, graceful performance degradation.

\begin{table}[!t]
  \centering
  \footnotesize
  \renewcommand{\arraystretch}{1.5}
  \caption{The target confusion matrix obtained in our experiments (shown is error rate in \%). Our algorithm made few errors (wrong target assignments) and the few that were made can be seen to correspond to genuinely similar objects. }
  \vspace{3pt}
  \begin{tabular}{l||cccccc|c}
     \Hline
                 &   Car 1 &   Tent &   Van 1 &   Car 2 &   Van 2 &   Car 3 &   Total \\
     \hline
      Car 1 &    --   &   0.0  &   3.4   &   0.0   &   0.0   &   4.7   &   7.0\\
      Tent  &    0.0  &   --   &   0.0   &   0.0   &   0.0   &   0.0   &   0.0 \\
      Van 1 &    2.2  &   0.0  &   --    &   0.0   &   0.0   &   1.5   &   3.7 \\
      Car 2 &    0.0  &   0.0  &   0.0   &   --    &   1.9   &   0.0   &   1.9 \\
      Van 2 &    0.0  &   0.0  &   0.0   &   2.9   &   --    &   0.0   &   2.9 \\
      Car 3 &    6.7  &   0.0  &   0.8   &   0.0   &   0.0   &   --    &   7.5 \\
     \Hline
  \end{tabular}
  \label{t:cm}
\end{table}

\section{Conclusions} This paper presented preliminary experiments and findings for target recognition from unmanned reconnaissance aircraft using a method based on canonical correlations, a well known statistical method for comparing sets of high dimensional vectors. A framework for employing canonical correlations in this scenario was described, followed by a description of experiments conduced to assess its effectiveness. These preliminary results show promising performance with high correct recognition rate and graceful performance decay in the presence of a reducing amount of data.

\subsection{Future work} Results reported here encourage and call for more experimental and research effort in the development of a canonical correlations-based target recognition system. These include:
\begin{itemize}
  \item \textbf{Data variation modelling:} In this paper we only evaluated the performance of canonical correlations using linear subspaces. This approach is almost always inferior to one which takes into account the nonlinear nature of object appearance manifolds. Specifically, we would like to investigate the performance of a method which would represent each 360$^{\circ}$ appearance variation as a set of subspaces, which are then mutually compared in a manner similar to that described in~\cite{AranCipo2006e}. Another promising direction involves the use of probabilistic extensions of canonical correlations~\cite{Aran2014}.
  \item \textbf{Different appearance representation:} Here we only explored the use of raw image appearance to model target appearance. The use of more complex representations, e.g.\ based on oriented gradients~\cite{Lowe2004,DalaTrig2005,Aran2012e}, robust edges~\cite{AranPhamVenk2015c,AranPhamVenk2015e,Aran2013} or colour invariants~\cite{Aran2012b}, could result in increased robustness of the method and possibly remove the need for synthetic data augmentation over full 360$^{\circ}$.
\end{itemize}

\bibliographystyle{ieee}
\bibliography{../../../my_bibliography}

\balance

\end{document}